\pdfoutput=1

\documentclass[11pt]{article}

\usepackage[]{ACL2023}

\usepackage{times}
\usepackage{latexsym}

\usepackage[T1]{fontenc}

\usepackage[utf8]{inputenc}

\usepackage{microtype}

\usepackage{inconsolata}

\usepackage{amsmath}
\usepackage{amssymb}
\usepackage{booktabs}
\usepackage{multirow}
\usepackage{graphicx}
\usepackage[ruled,vlined]{algorithm2e}
\usepackage{algorithmic}
\usepackage{color}
\usepackage{makecell}

\title{MORPHEUS\thanks{$^*$Morpheus excels at imitating others' forms, effortlessly embodying their physicality, posture, facial expressions, and vocal tone. He flawlessly embodies every intricate detail, including attire and unique mannerisms. --- \textit{Metamorphoses} by Ovid}: Modeling Role from Personalized Dialogue History by Exploring and Utilizing Latent Space}

\author{Yihong Tang$^{1}$, Bo Wang$^{2,}$\thanks{$^*$Corresponding author.}, Dongming Zhao$^3$ \\ {\bf Xiaojia Jin$^3$, Jijun Zhang$^3$, Ruifang He$^2$, Yuexian Hou$^2$} \\
        $^1$School of New Media and Communication, Tianjin University, Tianjin, China \\ 
        $^2$College of Intelligence and Computing, Tianjin University, Tianjin, China \\ 
        $^3$AI Lab, China Mobile Communication Group Tianjin Co., Ltd. \\
        \texttt{\{toyhom, bo\_wang\}@tju.edu.cn}
}

\begin{document}
\maketitle

\begin{abstract}

Personalized Dialogue Generation (PDG) aims to create coherent responses according to roles or personas.
Traditional PDG relies on external role data, which can be scarce and raise privacy concerns. Approaches address these issues by extracting role information from dialogue history, which often fail to generically model roles in continuous space.
To overcome these limitations, we introduce a novel framework \textbf{MO}dels \textbf{R}oles from \textbf{P}ersonalized Dialogue \textbf{H}istory by \textbf{E}xploring and \textbf{U}tilizing Latent \textbf{S}pace (MORPHEUS) through a three-stage training process. Specifically, we create a persona codebook to represent roles in latent space compactly, and this codebook is used to construct a posterior distribution of role information. This method enables the model to generalize across roles, allowing the generation of personalized dialogues even for unseen roles.
Experiments on both Chinese and English datasets demonstrate that MORPHEUS enhances the extraction of role information, and improves response generation without external role data. Additionally, MORPHEUS can be considered an efficient fine-tuning for large language models.
\end{abstract}

\section{Introduction}
Personalized Dialogue Generation \citep{Chen2024RecentTI} (PDG) aims to model role bases according to given roles or personas, thus generating role-relevant responses to meet users' individual needs. Many methods have achieved commendable results by explicitly incorporating external role data into prompts and capturing the direct relationship between dialogue and roles during inference. 

\begin{figure}[htbp]
\centering
\resizebox{.48\textwidth}{!}{
\includegraphics[width=.48\textwidth,keepaspectratio]{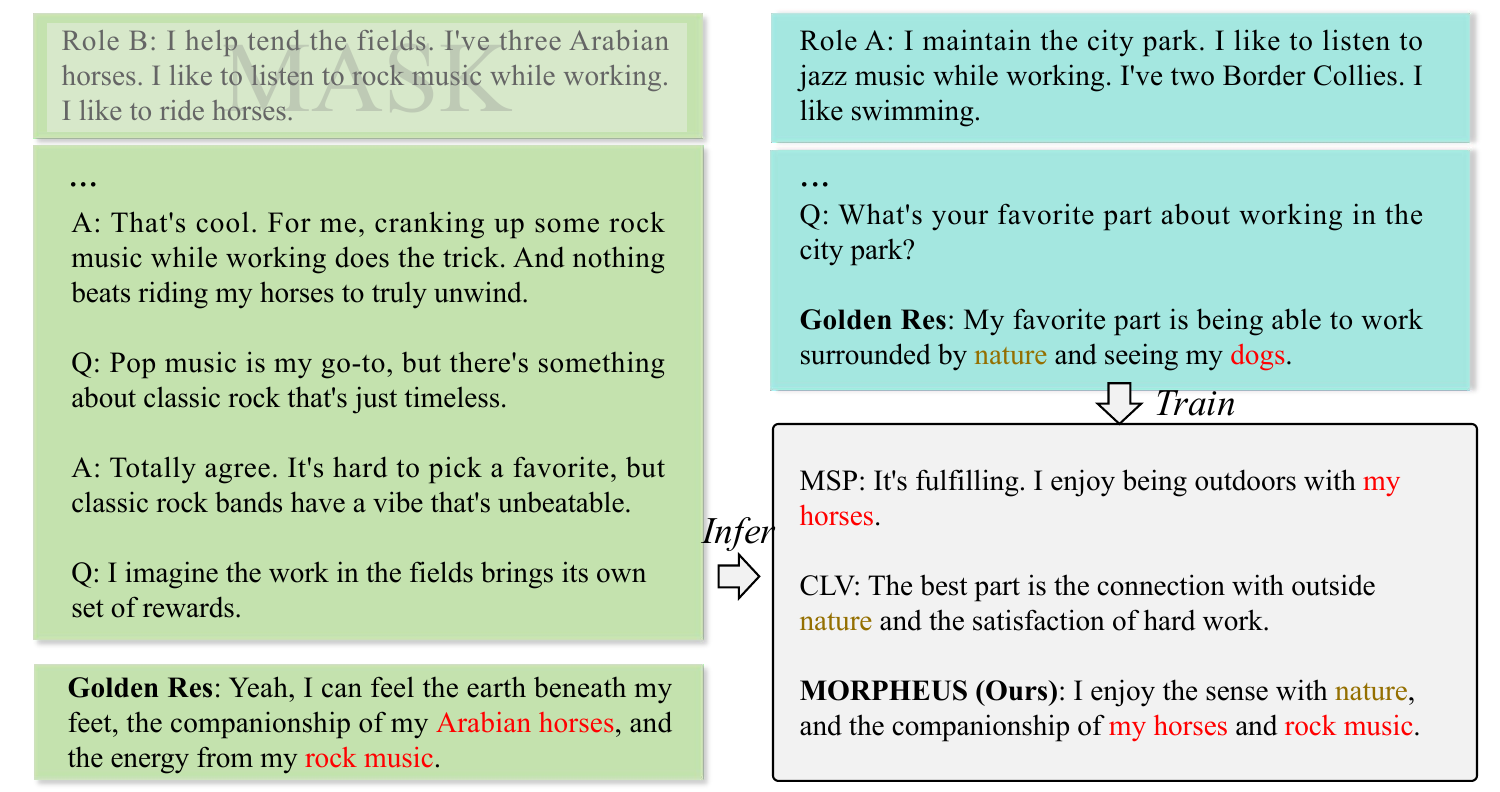}}
\caption{In the task of PDG when role data is masked, an example generated by two state-of-the-art (SOTA) models, MSP and CLV,  alongside our model MORPHEUS, is presented. Roles A and B are similar, and we expect models to generalize from the data about A to learn how to generate B's response.}
\label{fig:ex}
\vspace{-0.3cm}
\end{figure}

Despite their successes, existing PDG methods overly rely on extracting information from sparse external role knowledge bases~\citep{zhang-2018-personalizing,song2019exploiting,wolf-2019-TransferTransfo,liu-2020-impress,song-2021-bob}; or brief textual persona descriptions~\citep{Qian-2018-assign,Zheng-2020-Pre-Training,song-2021-bob}. However, such external role data is often lacking and may raise privacy concerns. 

Therefore, researchers attempt to leverage the role information presented in dialogue history~\citep{li-2016-persona, Chan2019ModelingPI, ma-2021-onechatbot}. The MSP~\citep{zhong-etal-2022-less} introduces a token-level retrieval-generation system that retrieves relevant personalized dialogues from a corpus and extracts tokens as prompts for generating responses. While MSP employs a retrieval strategy for the implicit modeling of roles, it treats different roles in isolation and overlooks the inherent associations across roles. As illustrated in Figure \ref{fig:ex}, these models must grasp the concept that, similar to Role A, Role B derives comfort and enjoyment from ``nature'' and certain aspects of personal work and life, such as listening to music and spending time with pets, and utilize this understanding to craft appropriate responses, even when the persona data of Role B is masked. 

To address this issue, CLV~\citep{Tang2023EnhancingPD} attempts to model roles in latent space, dealing with similar characteristics through implicit clustering. Furthermore, PersonaPKL~\citep{Han2023PersonaPKTBP} trains the model to perceive roles and then transfers this role perception ability to specific roles. However, these methods focus only on the associations between local role features, which leads to the model's difficulty in generating personalized responses for unseen roles. In fact, by learning dialogues based on roles, models could potentially acquire a universal ability to model unseen roles.

To achieve this insight, we introduce MORPHEUS, a framework that \textbf{MO}dels \textbf{R}oles from \textbf{P}ersonalized Dialogue \textbf{H}istory by \textbf{E}xploring and \textbf{U}tilizing latent \textbf{S}pace. First, we train the model to perceive roles, where it is supervised to generate personalized responses based on personas data and history, thereby establishing the relationship between persona information and dialogue history. Next, we create a persona codebook to represent roles compactly in latent space, initializing it with the encoded representations of existing personas. Finally, we construct a posterior distribution of persona information in the codebook using explicit personas and predict these codes using the dialogue history. Ultimately, our three-stage training method results in a codebook that generalizes the representation of all roles. During inference, MORPHEUS can automatically infer codes from the dialogue history, extract the code representations from the codebook, and use them to generate dialogue.

In summary, our contributions are as follows:
(1) To our knowledge, this is the first systematic work considering role generalization and the generation of personalized dialogues for unseen roles;
(2) We propose a novel framework MORPHEUS that improves the ability to generalize roles in the face of unseen roles or no external role data by modeling roles globally in the latent space;
(3) By applying MORPHEUS, experimental results show significant improvements in the personalization of dialogue generation. Additionally, MORPHEUS can serve as a parameters-efficient fine-tuning approach for large language models.

\begin{figure*}[htbp]
\centering
\includegraphics[height=.33\textheight,width=.98\textwidth]{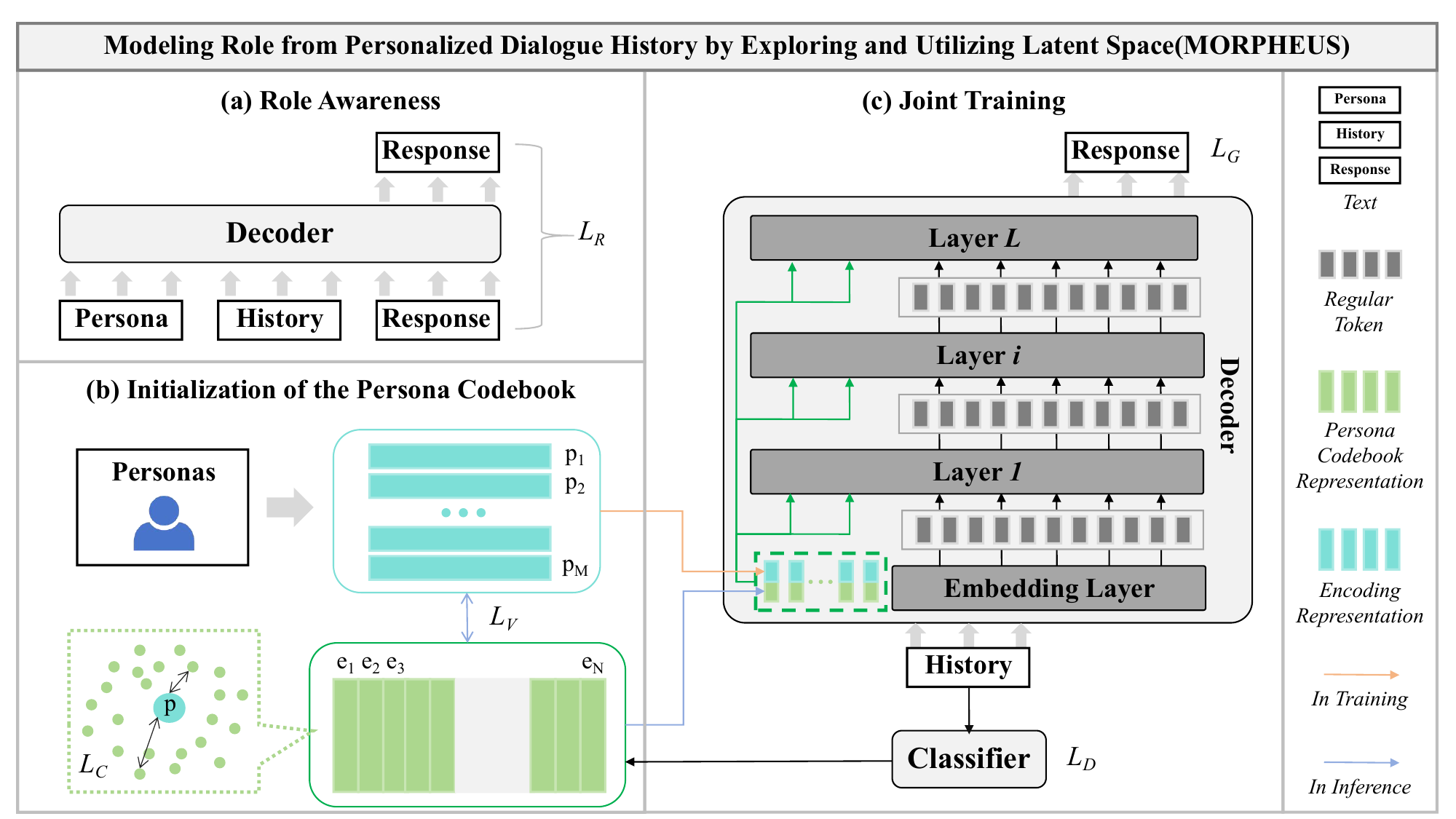}
\caption{The overview structure of the proposed model.}
\label{fig:overview}
\vspace{-0.2cm}
\end{figure*}

\section{Related Work}

\paragraph{Personalized Dialogue Generation}
Previous research has demonstrated that incorporating personality traits into dialogue generation is crucial~\citep{Chen2024FromPT} for personalizing dialogue systems. The question of how to maximize the utilization of explicit persona data when available is a concern for most studies. These data include factual role descriptions~\citep{Qian-2018-assign,Zheng-2020-Pre-Training,song-2021-bob} (e.g., "I enjoy swimming"), personality traits (e.g., introverted), knowledge bases~\citep{zhang-2018-personalizing,song2019exploiting,wolf-2019-TransferTransfo,liu-2020-impress,song-2021-bob} (e.g., gender: male), and speaking styles~\citep{Lu2023MIRACLETP} (e.g., excited). 
In a word, the goal of all research is to model a role profile with personality, engaging in personalized, continuous dialogue with users. However, compared to explicit data, which is scarce and difficult to extract from the real world, dialogue histories can be collected on a large scale and also contain rich persona information. Modeling personality solely from dialogue history is challenging but offers significant advantages, including avoiding privacy issues associated with explicit data. 

The essence of modeling personalities from dialogue history is to extract and utilize generic persona information. Some past studies, such as DHAP~\citep{ma-2021-onechatbot} and MSP~\citep{Tang2023EnhancingPD}, enhance current dialogue generation by explicitly retrieving relevant dialogues from the entire corpus. These approaches achieve the goal of modeling roles from dialogue history but suffer from the limitations of corpus size and the inefficiency of retrieval-based methods. Unlike retrieve-based studies, such as PersonaWAE~\citep{Chan2019ModelingPI}, model personalization in the latent space for the user side without considering the personality of the respondent. PersonalPKT~\citep{Han2023PersonaPKTBP}, achieve the goal of inferring dialogue for target roles without explicit data by efficiently fine-tuning the model in two stages with explicit persona data, but this still requires additional training for each target. CLV~\citep{Tang2023EnhancingPD} argues that explicit persona descriptions hide certain specific angle information. It clusters persona information in a latent variable space and obtains information solely from dialogue history through a CAVE~ \citep{Zhao2017LearningDD}. Furthermore, we believe that all persona information can actually be compressed in a larger latent space for generalization, which can then be utilized during dialogue generation.

\section{Methodology}

\subsection{Overview}
In this section, we first formalize the personalized dialogue generation with masked explicit person data and provide an overview of the framework we propose. Given a set of roles $U$, for each role $u_i$ in $U$, $\mathcal{P}$ represents the textual persona data to that role. Subsequently, there exists a multi-turn dialogue history $\mathcal C$ between $u_i$ and $u_j$, with the current turn awaiting a response from $u_i$. Our objective is to generate a personalized response $\mathcal R$ that aligns closely with the $u_i$ in response to the dialogue history $\mathcal C$. As previously mentioned, modeling persona or role profiles directly from dialogue history signifies that, during inference, the model does not require external data $\mathcal P_{u_i}$. Instead, it leverages the persona information $\mathcal P$ implicit within the $\mathcal C$: 
\begin{align}
P(\mathcal{R}|\mathcal{C}) = \sum_{\mathcal{P}} P(\mathcal{R}|\mathcal{C},\mathcal{P}) \cdot P(\mathcal{P}|\mathcal{C}).
\end{align}
As depicted in Figure \ref{fig:overview}, our proposed framework is divided into three primary training stages: (1) Role Awareness. Typically, an unconditional decoder is trained on a broad corpus, lacking specific perception capabilities for dialogues and persona information. The goal of this stage is to fine-tune the model to enhance its perception of persona and the relationship between persona and dialogue, thereby enabling the generation of dialogues that are more aligned with role traits. (2) Persona Codebook(PC) Initialization. Considering our model will directly model roles from dialogue history and PC, the existing codebook might not align with the encoding representations of personas. Hence, we explored various initialization methods for PC, finding that PC generalized from specific persona encoding representations significantly outperforms random initialization in effectiveness. (3) Joint Training. This crucial training stage involves the model taking dialogue history as input, sampling PC representations from PC based on the implicit persona information within the dialogue, thereby producing personalized responses.

\subsection{Role Awareness}
In decoder-only architecture models, it is common practice to concatenate the persona with the dialogue history and then generate responses in an auto-regressive manner. Similarly, the hidden representation of role data can also be directly concatenated at the beginning of the model in the form of past keys and values. Drawing on the work of ~\citet{Han2023PersonaPKTBP}, we found that forcing the model to generate responses based on dialogue history, conditioned on the hidden state form of role data, can enhance the model's joint perceptual ability of persona information and dialogue history. Following this approach, a persona segment about the role is encoded, called encoding representation $p_i$, and then only the keys and values are concatenated at the forefront of the model. Subsequently, we fine-tune the model:
\begin{equation}\label{formula:roleloss}
\begin{aligned}
L_{R} = \sum_{j}^{l}\log{P(\hat{\mathcal R_i}|\mathcal P,\mathcal C,\mathcal R_{<j})} .
\end{aligned}
\end{equation}
\subsection{Initialization of the Persona Codebook}
Upon creating a Persona Codebook(PC) $e \in \mathbb{R}^{N\times d}$ where N is the size of the discrete latent space and $d$ is the hidden size. The $e$ contains $N$ PC representation $e_k$.  Our aim is to compactly utilize the PC to represent a wide variety of persona information within this latent space. With the model already having a strong perception of roles, a critical issue arises: whether the $e_k$ in the PC can align with the representation $p_i$ of textual persona data encoded:
\begin{align}
\mathcal P_0,\mathcal P_2,...,\mathcal P_{M-1} = Split(\mathcal P), \\
p_i = Encoder(\mathcal P_i),
\end{align}
where $p_i \in \mathbb{R}^{d}$. $Split$ represents an explicit split at the string level, with a period as the separator, and $M$ comes from the theory in CLV~\citep{Tang2023EnhancingPD} that the description angle in the persona data can be divided into $M$ categories.

This alignment is crucial for the $e_k$ to be effective during decoding. Indeed, we find that the initialization of the PC is paramount in addressing this issue. Utilizing random uniform distribution for initializing often results in difficulty enhancing PC during subsequent parameter updates, potentially leading to the failure of joint training. Conversely, a well-initialized PC significantly smoothens the joint training process described in Section \ref{j_t}. To this end, we design three novel initialization methods to tackle this challenge: sequential persona initialization, average persona initialization, and EM~\citep{Dempster1977MaximumLF} (Expectation-Maximization) initialization.

During training, sequential initialization populates the uninitialized portions of the PC with the encoding representations $p_i$ of persona data from the current batch in sequence, while average initialization performs an additional averaging operation on the batch's $p_i$. The EM algorithm-based initialization emerged as the most effective, operating independently from the joint training process. In EM initialization, we assume that the $p_i$ of persona information is sampled from one of several normal distributions, each centered at a $e_k$. The means and variances of these distributions are treated as parameters to be estimated by the EM algorithm.

\textbf{The Expectation Step}
For a PC of size N, we establish N normal distributions, using all persona data encoding representations from the training set as samples, totaling $|D|$. For each encoding representation \(p_i\), we calculate its posterior probability \(P(z_{i}=k|p_i)\) of belonging to each normal distribution \(N_k\) (with mean \(\mu_k\) and variance \(\sigma_k^2\)), where \(z_{i}=k\) denotes the latent variable indicating that \(p_i\) belongs to the \(k\)-th distribution:

\begin{equation}
\small
\begin{aligned}
P(z_{i}=k|p_i) = \frac{P(p_i|z_{i}=k)P(z_{i}=k)}{\sum_{j=1}^{N}P(p_i|z_{i}=j)P(z_{i}=j)},
\end{aligned}
\end{equation}

where \(P(z_{i}=k)\) is the prior probability of the latent variable \(z_{i}=k\), assumed to be uniform distribution.

\textbf{The Maximization Step}
The goal during the M-step is to update the parameters (mean \(\mu_k\) and variance \(\sigma_k^2\)) of each distribution to maximize the expected log-likelihood of the posterior probabilities calculated in the current step:
\begin{align}
\mu_k = \frac{\sum_{i=1}^{|D|}P(z_{i}=k|p_i)p_i}{\sum_{i=1}^{|D|}P(z_{i}=k|p_i)}
\end{align}
\begin{align}
\sigma_k^2 = \frac{\sum_{i=1}^{|D|}P(z_{i}=k|p_i)(p_i - \mu_k)^2}{\sum_{i=1}^{|D|}P(z_{i}=k|p_i)}
\end{align}
The denominator \(\sum_{i=1}^{|D|}P(z_{i}=k|p_i)\) represents the weighted count of all encoding representations \(p_i\) belonging to distribution \(k\), essentially the "effective" sample size considering the uncertainty of each sample belonging to each code $z$. Through this method, we obtain accurate parameters for the normal distributions, using the derived mean $\mu_k$ as the initial value $e_k$ for the $z_k$.

\subsection{Joint Training}\label{j_t}
Following the initial setup of both components, our model embarks on learning how to predict codes based on the implicit persona information contained within the dialogue history. It then extracts PC representation $e_k$ from the Persona Codebook (PC), thereby generating personalized responses. To train this mechanism, our joint training regimen is bifurcated into two concurrent steps: (1) Code Index Prediction and (2) Training of the PC.

During the training phase of the PC, following the approach of VQ-VAE \citep{Oord2017NeuralDR}, we identify the $e_k$ that most closely aligns with the persona's encoding representation $p_i$:
\begin{align}
k = \underset{k}{\arg\min} \left \| p_i - e_k \right \|_2 .
\end{align}
We optimize the proximity of these PC representations by minimizing their distance, employing the following loss function:
\begin{equation}\label{formula:vaeloss}
\begin{aligned}
L_{V} = \left \| sg\left [ p_i \right ] - e_k  \right \|_2^2 + \beta \left \| sg\left [ e_k \right ] - p_i  \right \|_2^2 ,
\end{aligned}
\end{equation}
where the \(sg\) denotes the stop gradient operator, defined as an identity operation during forward computation but possesses a zero partial derivative, effectively rendering its operands as un-updated constants. 
The encoder optimizes the last loss term, while the embeddings are optimized by the first loss term. 
Our findings indicate that the algorithm exhibits sensitivity towards \(\beta\) because the encoder's parameter volume significantly exceeds that of the intermediary processes. Across all experiments, we utilize \(\beta = 0.05\).

Subsequently, the encoded representations participate in autoregressive dialogue generation as past keys and values, with its loss function as:
\begin{align}
p_{G} = concat(p_{1},p_{2},...,p_{M}) ,\\
L_G = -\sum_{j}^{l}\log{P(\mathcal R_j|p_{G}, \mathcal C,\mathcal R_{<j})} , \label{formula:genloss}
\end{align}
In the realm of Code Index Prediction, the code index $k$ that is nearest to the $p_i$ is employed as the label. The dialogue history's hidden states $c$ are inputs, utilizing a multilayer MLP as the classifier:
\begin{align}
c &= Decoder(\mathcal C) ,\\
L_D &= -\log{P(y=k|c)} ,
\end{align}
Furthermore, to ensure the PC possesses sufficient diversity and generalization capability, we introduce additional contrastive learning \citep{Gao2021SimCSESC} loss pertaining to each latent variable within the codebook. This measure is aimed at enhancing the representational breadth and applicability of the PC across varied dialogic scenarios:
\begin{align}
L_C &= -log\frac{e^{sim(p_i,e_k)/\tau}}{\sum_{j=1}^{N}e^{sim(p_i,e_j)/\tau}}, 
\end{align}
where $\tau$ is a temperature hyperparameter and $sim(p_i,e_i)$ is the cosine similarity.

\begin{table}[t]
\centering
\resizebox{.93\columnwidth}{!}{
\begin{tabular}{llll}
\toprule
\multicolumn{1}{c}{\textbf{Dataset}} &
  \multicolumn{1}{c}{\textbf{\# Train}} &
  \multicolumn{1}{c}{\textbf{\# Valid}} &
  \multicolumn{1}{c}{\textbf{\# Test}} \\ \midrule
ConvAI2     & 43,410           & 4,213            & 2,138            \\
Baidu PersonaChat    & 376,016         & 19,923 &  4,456    \\ \bottomrule
\end{tabular}}
\caption{Statistics of persona dialogue datasets.}
\label{tab:datasets}
\vspace{-0.5cm}
\end{table}

\begin{table*}[t!]
\centering
\small
\resizebox{\textwidth}{!}{
\begin{tabular}{llrrrrrrrr}
\toprule
&  & \multicolumn{3}{c}{Coherence} & \multicolumn{3}{c}{Diversity} & \multicolumn{2}{c}{Consistency} \\  
\cmidrule(lr){3-5}\cmidrule(lr){6-8}\cmidrule(lr){9-10}  
& & {BLEU-1} & {ROUGE-L} & {Coh.Score} &  {Dist-1} & {Dist-2} & {sBLEU} $\downarrow$ &  {Coh-Con.Score} &  {P-Co} \\ 
\midrule

\multirow{7}[0]{*}{ConvAI2} &  Seq2Seq & 3.41  & 5.48  & 35.89  & 2.04  & 3.98  & 13.22  & 10.85  & 3.13  \\ 
        & GPT-2 & 6.50  & 10.91  & 58.79  & 4.71  & 25.28  & 10.51  & 13.29  & 4.59  \\ 
        & DHAP & 7.37  & 9.97  & 63.21  & 5.60  & \textbf{29.81}  & 9.85  & 16.04  & 9.27  \\ 
        & MSP & 8.55  & 10.96  & 64.19  & 5.11  & 28.60  & 9.93  & 15.45  & 8.92  \\ 
        & PersonaPKT & 8.70  & 11.08  & 60.58  & \textbf{6.30}  & 26.72  & 9.34  & 24.87  & 9.26  \\ 
        & CLV & 11.16  & 15.04  & 70.83  & 4.31  & 26.17  & 10.14  & 23.01  & 9.38  \\ 
        & \textbf{MORPHEUS(Ours)} & \textbf{12.67}  & \textbf{16.18}  & \textbf{73.19}  & 5.89  & 28.74  & \textbf{8.97}  & \textbf{31.57}  & \textbf{11.64}  \\ 
\midrule
\multirow{7}[0]{*}{\makecell{Baidu \\ PersonaChat}} &

        Seq2Seq & 7.98  & 8.24  & 40.11  & 0.97  & 5.19  & 16.79  & 8.96  & 3.11  \\ 
        & GPT-2 & 10.16  & 12.29  & 49.72  & 3.08  & 20.98  & 13.32  & 12.14  & 5.30  \\ 
        & DHAP & 11.23  & 11.58  & 53.89  & 3.11  & 22.10  & 13.42  & 12.30  & 7.95  \\ 
        & MSP & 14.44  & 13.24  & 58.59  & \textbf{3.37}  & 22.41  & 13.95  & 14.37  & 8.23  \\ 
        & PersonaPKT & 13.82  & 15.57  & 53.95  & 2.98  & 21.83  & 13.10  & 19.86  & 8.17  \\ 
        & CLV & 18.77  & 21.82  & 59.74  & 2.42  & 21.17  & 14.57  & 18.15  & 8.39  \\ 
        & \textbf{MORPHEUS(Ours)} & \textbf{19.70}  & \textbf{24.64}  & \textbf{62.45}  & 3.07  & \textbf{23.05}  & \textbf{12.56}  & \textbf{29.93}  & \textbf{10.96} \\

\bottomrule
\end{tabular}}
\caption{Automatic evaluation on two datasets. The best results are in \textbf{bold}.}\label{tab:auto}
\vspace{-0.3cm}
\end{table*}

\section{Experiments}

\subsection{Datasets}
\textbf{ConvAI2}~\citep{dinan2019second} is an enriched English dataset centered on personalized dialogues that stem from specific role profiles. This dataset, an extension of the original PersonaChat~\citep{zhang-etal-2018-personalizing}, was meticulously curated by crowd workers to enhance its quality and relevance.

In a parallel vein, \textbf{Baidu PersonaChat}\footnote{\url{https://www.luge.ai/\#/luge/dataDetail?id=38}}, a Chinese counterpart to ConvAI2, mirrors its structure by emphasizing personalization and leveraging individual details to drive engaging dialogues.

Table~\ref{tab:datasets} presents statistics for the two datasets. We utilize these persona descriptions only during the training stage. We use the dataset curated by~\cite {Tang2023EnhancingPD} but ensure no role leaks for the training, validation, and test data.

\subsection{Baselines}
In the exploration of personalized dialogue generation models without external role data, research in this domain can generally be categorized into two main types: First, the approach based on explicit retrieval, which relies on fetching the most relevant response or tokens from a predefined or dynamically generated response pool according to the current dialogue history. Second, the approach is based on implicit modeling, where the personalization or roles are learned and generated directly within the model itself, without the necessity for an explicitly defined response or tokens. Furthermore, when considering non-personalized dialogue generation models, we can categorize the existing model baselines into three distinct classes:

\paragraph{Non-Personalized Approaches}
The \textbf{Seq2Seq} ~\citep{Sutskever-2014-sequence} enhanced with an attention mechanism~\citep{Luong2015EffectiveAT} exemplifies a key sequence modeling with context-focusing capabilities. The pre-trained \textbf{GPT-2}~\citep{radford2019language} is a cornerstone for dialogue creation upon fine-tuning with dialogue-specific datasets. 

\paragraph{Approachs based on Explicit Retrieval}
\textbf{DHAP}~\citep{ma-2021-onechatbot} capitalizes on historical interaction data to dynamically craft query-sensitive user profiles, employing a specialized decoder for tailored responses. In a similar vein, \textbf{MSP}~\citep{zhong-etal-2022-less} propels personalized dialogue generation forward by identifying conversations from users with similar profiles using User and Topic Refiners, and refining tokens for training via a Token Refiner, thus achieving superior performance in generating personalized dialogues without relying on predefined personas.

\paragraph{Approaches based on Implicit Modeling}
\textbf{PersonaPKT}~\citep{Han2023PersonaPKTBP} utilizes minimal additional parameters to embed persona-specific traits, leveraging lightweight tuning methods to transitively model roles in two phases via continuous vectors.
\textbf{CLV}~\citep{Tang2023EnhancingPD} categorize persona information in latent space to enhance personalized response generation with dialogue history.

\paragraph{Implementation Details}
are in Appendix~\ref{sec:dps}.

\subsection{Evaluation Metrics}

In order to obtain accurate comparisons, we use both automatic and human evaluations.

\paragraph{Automatic Evaluation} 
We divide the automatic evaluation methods into three categories in order to evaluate and model the diversity, consistency, and coherence of the generated dialogues.

(1) \textbf{Coherence} BLEU-1/2~\citep{papineni-2002-bleu} and ROUGE-L~\citep{lin-2004-autorouge} are classical words overlap-based metrics for measuring the similarity between generated responses and factual responses, which we believe can indirectly measure the coherence of dialogues. According to \citet{Tang2023EnhancingPD}, BLEU-3/4 imposes overly strict coverage requirements, making it unable to reflect the coherence between dialogue history and the response. Therefore, we do not consider it as part of the coherence assessment.

(2) \textbf{Diversity}: The Dist-1/2~\citep{li-2016-diversity} assesses the presence of unique uni- or bi-grams within the generated responses, commonly employed for evaluating diversity at the corpus level. The self-BLEU (sBLEU)~\citep{Liu2022ComposableTC} computes the diversity by BLEU from the similarity between responses.

(3) \textbf{Consistency}: In dialogue generation, maintaining consistency between the generated responses and the role is crucial. Following \citet{Tang2023EnhancingPD}, we introduce two scoring models, Con.Score and Coh-Con.Score, to measure the consistency between responses and persona information. Furthermore, we employ the P-Co (Persona Cosine Comparison)~\citep{Song2019ExploitingPI}  as a supplementary IDF-weighted words overlap-based metric to assess the semantic similarity between model responses and personas.

\paragraph{Human Evaluation} 
Considering the auto evaluation criteria's uncertainty, we conduct human assessments for all models. Specifically, We extract 100 data points (histories, responses, and role data) and engage three well-educated annotators to rank the responses from various models and normalize them into specific scores ranging from 0 to 1 to save costs. Our evaluation centers on readability, diversity, consistency, and coherence, prompting evaluators to rank $6$ options for the five excellent model-generated responses and the factual responses. Detailed evaluation criteria can be found in Figure \ref{fig: human_crite} in Appendix \ref{app: human_crite}.

\subsection{Experimental Results}

\paragraph{Automatic Evaluation} 
Table \ref{tab:auto} presents the performance of all models on automatic metrics in both Chinese and English datasets for the average value obtained by three random seeds, allowing for a clear observation of the enhancements in key metrics by our MORPHEUS. Specifically: 

(1)\textbf{Coherence}: Whether through coverage metrics such as BLEU-1, Rouge-L, or learning indices like Coh-Con.Score, the MORPHEUS's performance in consistency stands out. 

(2)\textbf{Diversity}: MORPHEUS also excels at generating more diverse responses compared to other models. Additionally, we overcome the drawback commonly associated with latent variable models like CLV~\citep{Tang2023EnhancingPD}, which sacrifice diversity to achieve improvements in other performance metrics. 
This also indicates that MORPHEUS's efforts are closer to the intended modeling target, leading to the convergence in the meanings of the responses. We will analyze this reason further in Section~\ref{sec:as}.

(3) \textbf{Consistency}: The guidance from Con.Score indicates the model's ability to integrate persona information into generation, particularly evident in scenarios without external persona data, further highlighting the superiority of MORPHEUS. Experimental results demonstrate that the proposed model can generate more personalized responses than all baselines.

\paragraph{Human Evaluation} 
The human evaluation outcomes for ConvAI2 are depicted in Table~\ref{tab:human}. We computed the Fleiss Kappa coefficient~\citep{landis1977measurement} among the three annotators by a novel approach wherein the ordinal rankings are conceptualized as discrete categories. The Kappa value is $0.65$, indicating \textit{substantial agreement} among the annotators. Overall, the human annotations align with the automatic evaluation findings, showcasing the strengths of MORPHEUS in personalized dialogue generation and fundamental readability.

\begin{table}[t!]
\centering
\small
\resizebox{\columnwidth}{!}{
\begin{tabular}{lrrrr}
\toprule
Model & Readability & Diversity & Consistency & Coherence \\ \midrule

DHAP  & 0.71 & 0.72 & 0.69 & 0.71  \\ 
MSP   &  0.70 & 0.73 & 0.69 & 0.8  \\ 
PersonaPKT  &   0.63 & 0.65 & 0.61 & 0.69  \\ 
CLV  &  0.79 & 0.76 & 0.73 & 0.76 \\ 
\textbf{MORPHEUS(Ours)} &        \textbf{0.82} & \textbf{0.84} & \textbf{0.77} & \textbf{0.85}  \\ \midrule
Ground-Truth &  0.92 & 0.89 & 0.93 & 0.91  \\ 

\bottomrule

\end{tabular}}
\caption{The result of human evaluation on ConvAI2. 
}
\vspace{-0.2cm}
\label{tab:human}
\end{table}


\section{Further Analysis}\label{sec:fa}

Further analyses are based on the ConvAI2 dataset, and similar phenomena can be observed on Baidu PersonaChat.

\begin{table}[t!]
\centering
\small
\resizebox{\columnwidth}{!}{
\begin{tabular}{lrrrr}
\toprule
Models & BLEU-1 & Dist-1/2 & Coh-Con.Score & P-Co \\ \midrule


        MORPHEUS(Ours) & 12.67  & 5.89/28.74 & 31.57  & 11.64  \\ 
        \quad\textit{w/o} PC & 6.41  & 4.83/25.97 & 13.20  & 4.46  \\ 
        \quad\textit{w/o} CL & 9.24 & 4.19/24.93 & 23.02 & 8.96 \\ 
        \quad\textit{w/o} RA & 8.56 & 5.03/27.39 & 29.05 & 9.56 \\
        \midrule
        \quad\textit{i/w} Random & 10.99 & 5.12/27.95 & 28.53 & 10.32 \\ 
        \quad\textit{i/w} Se. & 11.14 & 5.55/28.14 & 28.15 & 10.94 \\ 
        \quad\textit{i/w} Mean & 11.69 & 5.41/28.30 & 30.12 & 11.19 \\ 

\bottomrule
\end{tabular}}
\caption{Ablation experiments results on ConvAI2.}
\label{tab:as}
\vspace{-0.5cm}
\end{table}

\begin{figure}[t!]
    \centering
    \includegraphics[width=\linewidth]{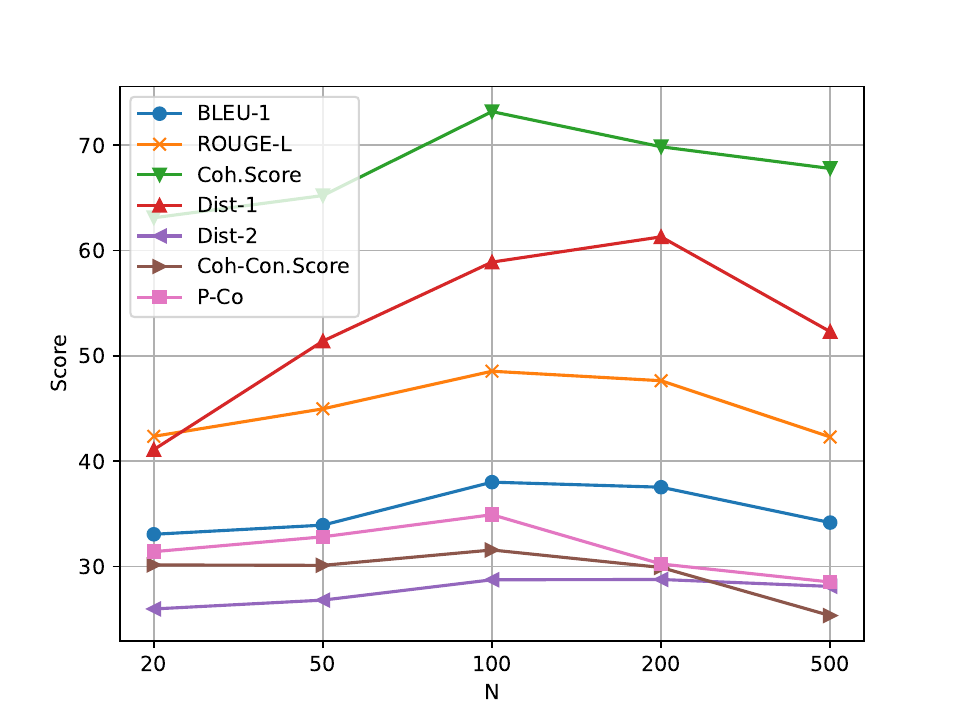}
    \caption[Caption for LOF]{Experiments with the different N on the ConvAI2 dataset. For ease of viewing, BLEU-1, rouge-L, P-Co are magnified by a factor of three and Dist-1 by a factor of ten.}
    \label{fig:n}
\vspace{-0.5cm}
\end{figure}

\begin{table*}[t]
\centering
\scriptsize
\begin{tabular}{p{.10\linewidth}p{.22\linewidth}p{.28\linewidth}p{.28\linewidth}}
\toprule
Role Data  & I enjoy \textcolor{red}{hiking}. I've never been abroad. I'm a bank teller. I've never been to college. My favorite phone is the iPhone. & I like watching cooking shows. I'm not good at swimming at all. I enjoy roller coasters and skydiving. I'm a vegetarian, and I like Hummers. & I help tend the fields. I've three Arabian horses. I like to listen to rock music while working. I like to ride \textcolor{red}{horses}. \\ \midrule
Dialogue History  & Q: Hi, what are you up to?
A: I'm getting ready to go out. How about you? What are you doing?

Q: I'm traveling with my girlfriend right now. & 
...Q: I'm a carpenter, I started my own business.
A: That's great!

Q: How old are you? I feel you are so young.
A: I just turned \textcolor{red}{17}.

Q: Oh, you're really young. what do you do? & Q: I spent my most of the time in study.
A: What are you studying?

Q: I am studying horses.
A: That's interesting.

Q: What is your favorite band?
A: I like rock.

Q: Me too. \\ \midrule 
Golden-Response & Do you like traveling everywhere? I do too. &  I don't have a job yet. I just moved here and haven't found one yet. & Do you like horses? I have three.  \\ \midrule
MSP & Could you tell me about some of your \textcolor{red}{travel experiences}? & I'm a farmer. & Do you like music? I like pop music.  \\ \midrule
CLV & I often \textcolor{red}{go out} with my girlfriend too. & I'm out running. & I don't like jazz.  \\ 
\midrule
MORPHEUS(Ours) & Do you like \textcolor{red}{traveling} everywhere? I do too. & I'm still a \textcolor{red}{student}. & Do you like \textcolor{red}{animals}?  \\
 \bottomrule
 \end{tabular}
 \caption{A case study. Keywords are marked in \textcolor{red}{red}.}
\label{tab:case study}
\vspace{-0.2cm}
\end{table*}

\subsection{Ablation Study} \label{sec:as}
We conducted ablation studies by removing specific modules of the MORPHEUS model, as presented in Table~\ref{tab:as}. Firstly, we examined the influence of the model's core mechanism—the persona codebook (PC)—on its performance; upon the removal of the PC, the model regresses to a basic GPT-2, and its performance aligns with that of GPT-2. If we merely eliminate the contrastive learning from the joint training while retaining the VQ-VAE loss function, there is a noticeable decline in the model's performance, especially the diversity. This underscores the effectiveness of employing contrastive learning to bring the encoding representations closer to the PC representations. Then, regarding various PC initialization methods, the EM(Ours) technique emerged as superior to random initial, sequential filling, or mean filling, suggesting that a global perspective on roles better optimizes the PC optimally. Finally, the removal of role-aware training results in a performance drop in certain metrics, indicating that aligning roles with dialogue through text is indispensable.

\subsection{Effects of size of the persona codebook} 
In MORPHEUS, persona information is compressed into the persona codebook (PC), where the size of the PC, denoted as $N$, has a significant impact on the performance as reported in Figure~\ref{fig:n}. When $N$ is small, the information compressed into the PC is more closely aligned with the generic information pertinent to the task of personalized dialogue generation rather than a specific role.
Conversely, too large $N$ can diminish the effectiveness of this compression. This reduction in effectiveness is attributed to an increase in noise resulting from an insufficient number of distinct personas, which leads to the dispersion of persona information and makes it challenging for the PC to be optimized effectively.
On balance, the best performance is observed when 
$N$ is $100$.

\vspace{-0.1cm}

\begin{table}[t!]
\centering
\resizebox{.98\columnwidth}{!}{
\begin{tabular}{lrrrrrrr}
        \toprule
        method & BLEU-1 & ROUGE-L & Dist-1 & Dist-2 & sBLEU $\downarrow$ & P-Co & TPS(\%) \\ 
        \midrule
        Prompt-Tuning & 8.15 & 10.02 & 4.24 & 11.27 & \textbf{7.62} & 8.81 & 0.03 \\ 
        Prefix-Tuning & 9.23 & 11.08 & 4.12 & 13.19 & 9.13 & 8.14 & 0.17 \\ 
        P-Tuning & 11.07 & 14.15 & 5.12 & 17.05 & 8.93 & 10.02 & 1.40 \\ 
        LoRA & 14.06 & 18.91 & 5.97 & 20.73 & 9.79 & 11.83 & 0.23 \\ 
        \textbf{MORPHEUS} & \textbf{16.98} & \textbf{20.03} & \textbf{7.34} & 32.28 & 7.92 & \textbf{12.30} & 0.16 \\ 
        \midrule
        LLaMA(Full) & 28.11 & 23.51 & 7.53 & 33.59 & 7.88 & 18.96 & 100 \\ 
\bottomrule
\end{tabular}}
\caption{Comparative experimental results of the PEFT methods based on LLaMA2-7B in ConvAI2. The TPS means trainable parameter size.}
\vspace{-0.5cm}
\label{tab:peft}
\end{table}

\subsection{PEFT}
Our approach of concatenating the persona codebook(PC) representation vectors to the forefront of the model is akin to prompt-tuning and p-tuning. Thus, our technique can be considered an efficient fine-tuning method. Specifically, we freeze the parameters of the decoder(LLaMA-2~\citep{Touvron2023LLaMAOA}) responsible for text generation and only optimize the parameters within the PC during training, with results presented in Table~\ref{tab:peft}. We compare our method against LoRA, prompt-tuning, p-tuning, and full-parameter fine-tuning under a similar parameter budget. Findings reveal that our method exhibits distinct superiority and yields results that are closer to those obtained from full parameter fine-tuning.

\subsection{Case Study}
For the case study, we select examples, including some that are clear failures, yet MORPHEUS consistently performs as well as or better than baselines. The results in Table \ref{tab:case study} indicate that MORPHEUS effectively aligns dialogue history and role. 

In case 1, the questioner is traveling, and while MSP and CLV both mentioned travel-related content, they failed to infer the respondent's hobby from the dialogue history. MORPHEUS accurately captured the fact that the respondent was preparing to go out, creating a response closely aligned with the role setting of "enjoy hiking". In case 2, when the questioner inquires about the profession, MSP and CLV both fail to notice the respondent's age mentioned in the history, whereas MORPHEUS generates a response that is consistent with the history and close to the ground-truth response. The same is true for case 3, due to the confusing history, MSP and CLV struggle to make relevant responses, while MORPHEUS's response is closer to the role.

Overall, facing previously unseen roles in the test set and without the external role data, MSP and CLV are easily misled by confusing history and unable to produce coherent responses. MORPHEUS overcomes this by accurately utilizing dialogue history to make reasonable inferences about the roles.

\section{Conclusion}
In this work, we introduce the MORPHEUS framework for generating personalized responses. Unlike existing efforts, we model roles in a generalized manner from the dialogue history of roles and align roles with dialogues through a multi-stage training process to enhance the personalization of dialogue generation. Experimental results demonstrate that our model is capable of generating highly personalized responses even when faced with previously unseen roles, efficiently utilizing persona information while mitigating the risk of privacy breaches.


\section*{Limitations}
The MORPHEUS framework introduces a significant advancement in the field of Personalized Dialogue Generation (PDG), particularly through its innovative use of latent space to model roles from dialogue history. However, like any study, there are areas that were beyond the scope of our current investigation. For instance, while MORPHEUS has been rigorously tested on Chinese and English datasets, its adaptability to other languages and dialects remains an area for future exploration. This limitation is not a reflection of the model's effectiveness but rather an opportunity to further validate its versatility across diverse linguistic contexts. Additionally, the current implementation of the persona codebook, although effective in compactly representing roles, offers potential for further refinement to enhance its representation capabilities. It's important to note that these considerations do not detract from the model's current achievements but highlight pathways for continued research and development. Furthermore, our study's focus on the technical development and application of MORPHEUS meant that a deeper dive into the broader societal implications of PDG technologies was outside our immediate scope. As the field progresses, these considerations will become increasingly important to address.

\section*{Ethics Statement}
This study acknowledges the ethical considerations inherent in the development and application of personalized dialogue generation technologies. While MORPHEUS aims to enhance the personalization of dialogue generation without relying on external role data, thus mitigating privacy concerns, it is imperative to consider the potential for misuse, such as the creation of misleading or harmful content under the guise of personalization. To address these concerns, the research strictly adheres to ethical guidelines, ensuring that all data used for training and evaluation is sourced from publicly available datasets that are free from unethical content. Moreover, the study employs anonymization techniques during human evaluation to protect the privacy of individuals. Despite these precautions, we recognize the importance of ongoing ethical scrutiny and the development of robust frameworks to prevent the misuse of personalized dialogue technologies, especially as they become more sophisticated and widely adopted.


\bibliography{anthology,custom}
\bibliographystyle{acl_natbib}

\appendix

\section{Appendix}
\label{sec:appendix}

\subsection{Default Parameter Settings}\label{sec:dps}
Our experiments are done based on pre-trained GPT-2, and we tried various model structures and hyperparameters, and the final hyperparameters are as follows: the size of GPT-2 embedding and GPT-2 hidden vector is 768. All word embedding dimensions are set to 768, and we use word2vec to initialize word embedding. The number of layers of Transformer is 12. The size of the persona codebook $N$ is set from 20 to 500(default is 100), the MLP input dimension and output dimension in the model are kept the same as the hidden vector, and the number of batches was set to 16. The maximum learning rate is 1e-4. The training of the proposed model was done on an Nvidia Telsa V100 16G GPU. The total training time takes approximately 10 hours. The temperature hyperparameter $\tau$ is $0.5$. The pre-trained models used in these experiments of this paper include gpt2\footnote{\url{https://huggingface.co/gpt2}}, gpt2-chinese-cluecorpussmall\footnote{\url{https://huggingface.co/uer/gpt2-chinese-cluecorpussmall}}, xlm-roberta-base\footnote{\url{https://huggingface.co/xlm-roberta-base}}, and chinese-roberta-wwm-ext\footnote{\url{https://huggingface.co/hfl/chinese-roberta-wwm-ext}}.

We use kernel sampling~\citep{Holtzman2020The} as our decoding strategy, use the Adam~\citep{Kingma-2014-adam} optimizer to train the model and use AdamW~\citep{loshchilov2018decoupled} to warm up the generator. Please refer to the published project for additional details, which is publicly available\footnote{\url{Comingsoon.}}.

When calculating BLEU, ROUGE, and Distinct, we use the TikToken-based word segmentation method \texttt{cl100k\_base}\footnote{\url{https://github.com/openai/tiktoken}} for all models to ensure cross-language data fairness.

\subsection{Evaluation Criteria of Human}\label{app: human_crite}
Upon the acquisition of human evaluators' rankings for the responses generated by six distinct models, the algorithm assigns scores in a manner that reflects both the quality and the relative positioning of these responses. Specifically, the scoring scheme is bifurcated based on whether a model's response precedes or follows the gold standard response in the ranking hierarchy.

For responses that are ranked higher than the gold standard, a uniform score of 1 is allocated. This decision is rooted in the assumption that surpassing the gold standard in human evaluators' rankings is indicative of exceptional quality, warranting the highest possible score within the algorithm's framework.

\begin{figure*}[htbp]
\centering
\includegraphics[width=.98\textwidth]{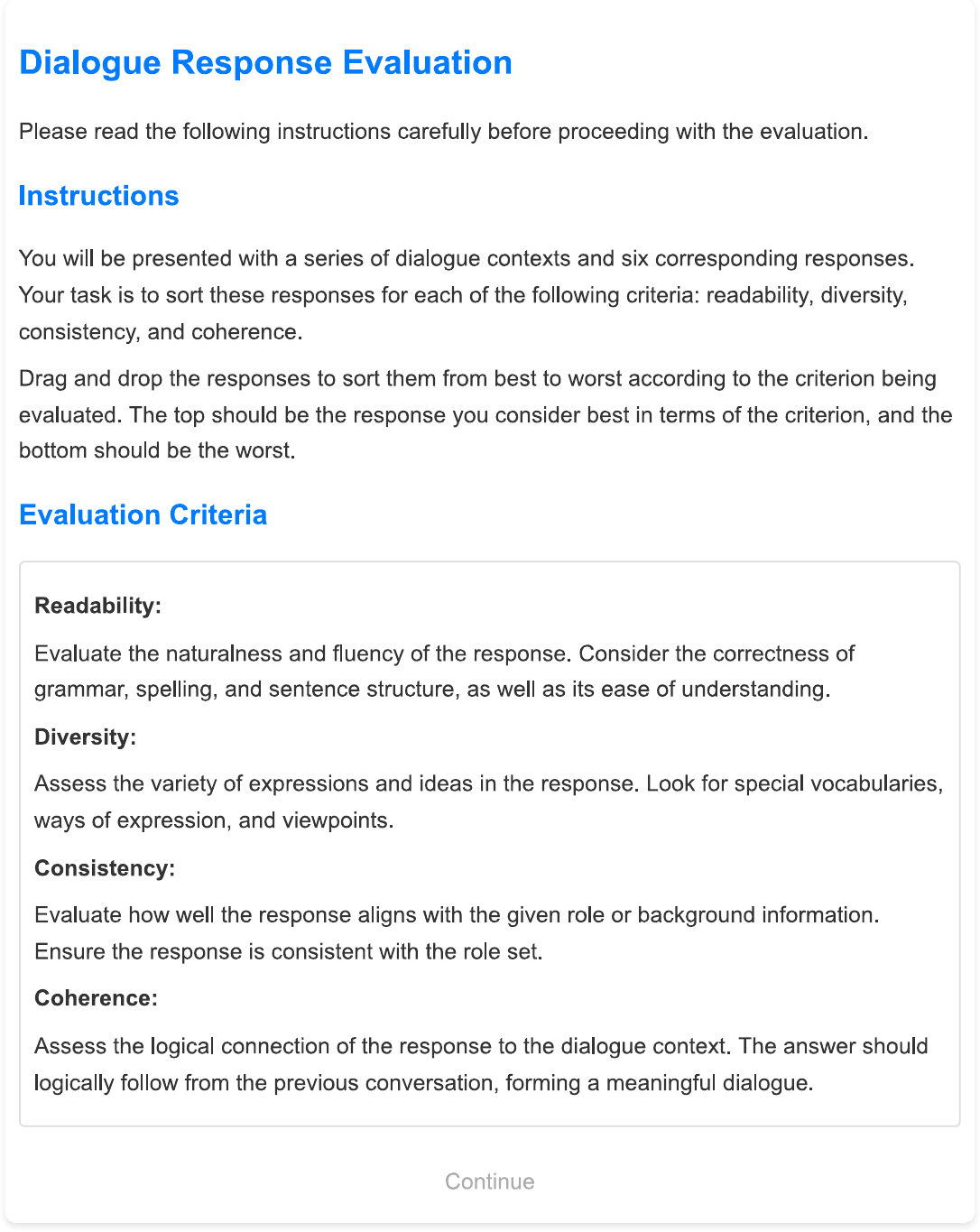}
\caption{The evaluation criteria of human.}
\label{fig: human_crite}
\end{figure*}

Moreover, recognizing the inherent challenge of applying the Fleiss Kappa coefficient directly to ranking-based assessments, a novel approach is adopted wherein the ordinal rankings provided by the evaluators are conceptualized as discrete categories. This conceptualization facilitates the application of the Fleiss Kappa coefficient to the task at hand, thereby enabling a quantifiable measure of inter-rater reliability in the context of a ranking-based evaluation framework.

we recruited 3 annotators with graduate degrees for \$20 an hour. Detailed evaluation criteria can be found in Figure \ref{fig: human_crite}.

\end{document}